# Soft Decision Tree classifier: explainable and extendable PyTorch implementation

Reuben R Shamir

KI Research Institute

## Background and overview

A soft decision tree (SDT) is a variant of the traditional decision tree where the splits at internal nodes are probabilistic rather than deterministic [1]. A single SDT that is trained for classification can approximate multiple hard decision trees. Therefore, it forms a compact representation for the trained tree-based classifier. It can be visualized and analysed to provide insights regarding the classifier results and the trained dataset (e.g. feature importance). Moreover, the SDT can be incorporated as a backbone for the development of novel tree-based machine-learning architectures. For a detailed description of SDT, the reader is referred to [1], [2].

We have implemented a SDT using PyTorch following Frosst et al. [2]. We developed a method to track the internal SDT parameters and to visualize them (Figure 1). Moreover, we introduce short-term memory SDT (SM-SDT) to demonstrate the extendibility of our implementation. SM-SDT is implemented by incorporating short-term memory capabilities into the nodes of the SDT. Specifically, each node is aware of the output of its immediate parent and grandparent in the tree. In addition, a neural linear layer was added at the input level of each node. Last, we have conducted simulation and clinical data experiments to validate and evaluate our implementation of SDT and SM-SDT. Accuracy results are reported below, the typical runtime of the SDT was 30-60 seconds on NVIDIA T4 GPU with 16 GB memory.

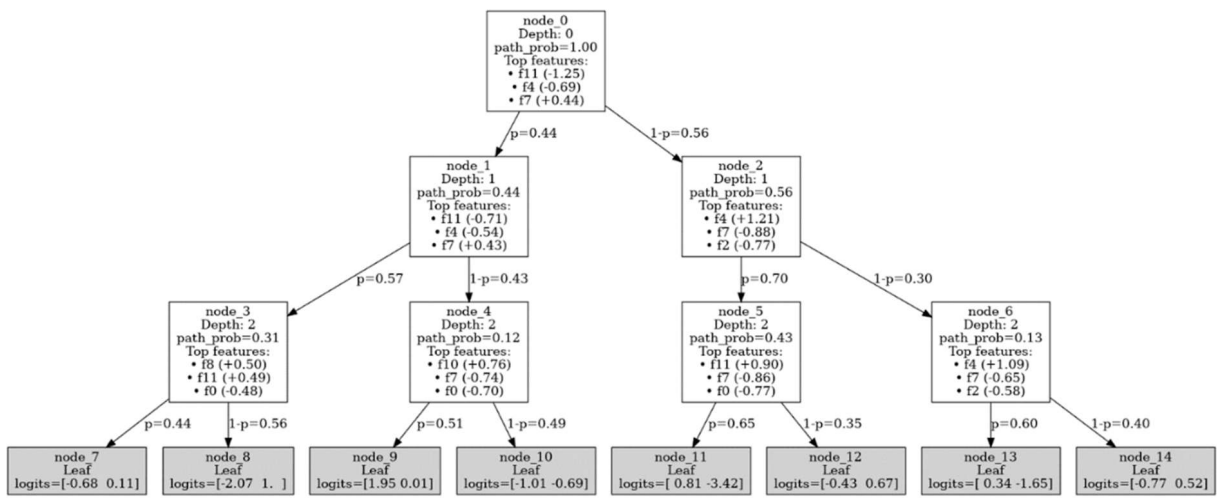

Figure 1: A visualization of a soft decision tree (SDT) classifier trained on heart failure dataset [3], [4]. The features with largest weights are presented at each internal node. The splits probabilities and leaf logits are presented as well. This information can be analysed and provide insights regarding the classifier results and the trained dataset (e.g. feature importance, classifier reasoning, etc.).

**Simulation experiments**

*Method:* We have incorporated the scikit-learn 'make_classification' module to generate random binary-outcome classification datasets with varying numbers of independent samples (1K, 100K, and 1M samples) and with varying numbers of features (50, 250, and 500 features). The number of features that contribute to the outcome remained 30 (constant) in all experiments. We have compared six classification algorithms for each simulated dataset: 1. Decision tree (DT); 2. logistic regression (LR); 3. a random forest with 1000 trees; 4. XGBoost; 5. SDT, and; 6. SM-SDT. All tree-based algorithms had a max depth of three.

*Results:* Figure 2 shows the average area under the curve (AUC) of five repetitive experiments using different random-generator seed values. XGBoost, SDT, and SM-SDT demonstrated similar AUC values and were associated with higher AUC values in comparison to the other three classification methods tested in this study, for most setups.

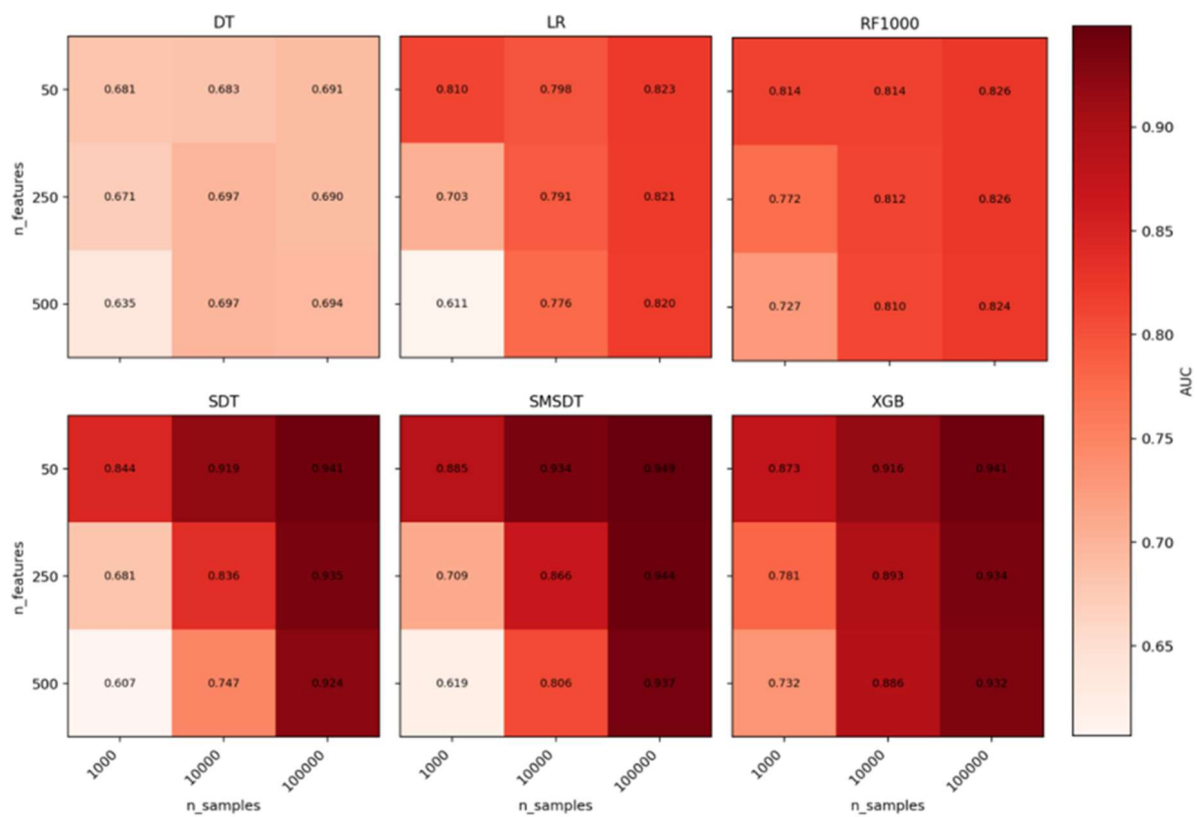

Figure 2: Average area under the curve (AUC) of six classification algorithms trained on simulated datasets of various sample- and feature-sizes.

**Clinical data experiments**

*Method:* We have downloaded, cleaned, and normalized seven clinical datasets. The outcomes were binarized in all cases. The seven datasets are: 1. Diagnostic Wisconsin breast cancer database [5]; 2. Heart disease (Cleveland subset) dataset [6]; 3. Heart failure dataset [3]; 4. Indian liver patient dataset (ILPD) [7]; 5. Pima Indians diabetes database (PIMA) [8]; 6. Stroke prediction dataset [9], and; 7. Thyroid disease dataset [10]. We have compared seven classification algorithms for each simulated dataset: 1. decision tree (DT); 2. logistic regression (LR); 3. a random forest with 100 trees; 4. a random forest with 1000 trees; 5. XGBoost; 6. SDT, and; 7. SM-SDT. All tree-based algorithms had a max depth of three. The train/test splits were 80/20% of the data in all cases. We have repeated this experiment five times with various random train/test splits.

*Results:* The average AUC values of all classification methods and clinical datasets are presented in Table 1. SDT and SM-SDT achieve state-of-the-art results, as the observed AUC is comparable to that of XGBoost, random forest, and logistic regression.

| Dataset \ Method | DT | LR | RF100 | RF1000 | SDT | SMSDT | XGB |
|---|---|---|---|---|---|---|---|
| Breast cancer | 0.93 | 0.99 | 0.98 | 0.98 | 0.99 | 0.99 | 0.99 |
| Heart cleveland | 0.8 | 0.91 | 0.91 | 0.91 | 0.9 | 0.9 | 0.87 |
| Heart failure | 0.84 | 0.87 | 0.92 | 0.92 | 0.86 | 0.86 | 0.9 |
| ILPD | 0.7 | 0.76 | 0.75 | 0.75 | 0.74 | 0.74 | 0.74 |
| Pima diabetes | 0.81 | 0.84 | 0.84 | 0.84 | 0.83 | 0.83 | 0.81 |
| Stroke | 0.8 | 0.83 | 0.83 | 0.83 | 0.82 | 0.82 | 0.79 |
| Thyroid sick | 0.96 | 0.96 | 0.98 | 0.98 | 0.98 | 0.98 | 1 |
| Average | 0.83 | 0.88 | 0.89 | 0.89 | 0.87 | 0.87 | 0.87 |

Table 1: Average AUC classification results. DT: Decision Tree; LR: Logistic Regression; RF100: Random Forest with 100 trees; RF1000: Random Forest with 1000 trees; SDT: Soft Decision Treel; XGB: XGBoost.

**Conclusions and future work**

We have developed and tested a PyTorch implementation of a soft-decision tree. We have developed a method to visualize the tree and expanded it with short-term memory capabilities. Our results suggest that the SDT achieves state-of-the-art AUC values. The drawbacks of the suggested method are that it requires a GPU and that the computation time is typically 30-60s in comparison to less than few (3-4) seconds observed for other methods. We have shared the code online on KI Research Institute GitHub webpage [11] and hope that others will be interested in further unveiling the practical potential of SDT.